\documentclass[runningheads]{llncs}
\usepackage[T1]{fontenc}
\usepackage{graphicx,verbatim}
\usepackage{subcaption}
\usepackage{hyperref}
\usepackage{color}

\begin{document}
\title{Big, Bright, or Invisible: A Frozen-Feature Benchmark of 3D CT Foundation Models}
\titlerunning{Frozen-Feature Benchmark of 3D CT Foundation Models}
%
\author{Maulik Chevli\inst{1}$^{\star}$ \and
Johannes Brandt\inst{1}$^{\star}$ \and
Rickmer Braren\inst{1,4} \and \\
Daniel Rueckert\inst{1,2,3} \and
Philip Müller\inst{1}}
\authorrunning{M. Chevli, J. Brandt et al.}
\institute{Chair for AI in Healthcare and Medicine, Technical University of Munich (TUM) and TUM University Hospital, Munich, Germany \and
Dept. of Computing, Imperial College London, UK \and
Munich Center for Machine Learning (MCML), Munich, Germany \and
Dept. of Diagnostic and Interventional Radiology, UKE Hamburg, Germany}

\maketitle              
\let\thefootnote\relax\footnotetext{$^{\star}$ These authors contributed equally.}
\begin{abstract}
Routine CT interpretation is inherently comprehensive, capturing incidental findings across the entire scan volume. 3D CT foundation models could assist this process by providing generalizable representations of anatomy and pathology. To evaluate their diagnostic breadth, we benchmark ten frozen CT encoders across three cohorts of thoracic CT scans, including an unseen internal clinical dataset, using $k$-nearest neighbors, zero-shot prompting, and linear probing. We find no universal state-of-the-art, with rankings fluctuating significantly depending on the evaluation context. While models combining fine-grained image tokenization with vision-language alignment generally perform best, a lightweight supervised encoder remains highly competitive, demonstrating that explicit labels can effectively substitute for scale. Crucially, rather than model architecture, we observe that the primary determinant of performance is a physical bottleneck: a finding's detectability scales with its contrast against surrounding tissue and its spatial extent. Through controlled within-organ comparisons, we empirically demonstrate that widespread or high-contrast abnormalities, such as devices and effusions, are reliably recovered. Conversely, small, low-contrast focal lesions remain a persistent challenge across all evaluated encoders. We attribute this to the inherent limitations of globally pooled embeddings, suggesting that accurately representing small, low-contrast structures will require region- or lesion-level pretraining.

\keywords{Foundation models \and Chest CT \and Incidental findings \and Multi-abnormality
classification \and Frozen features \and Transfer evaluation.}
\end{abstract}

\section{Introduction}
\begin{figure}[t]

\centering\includegraphics[width=0.8\textwidth]{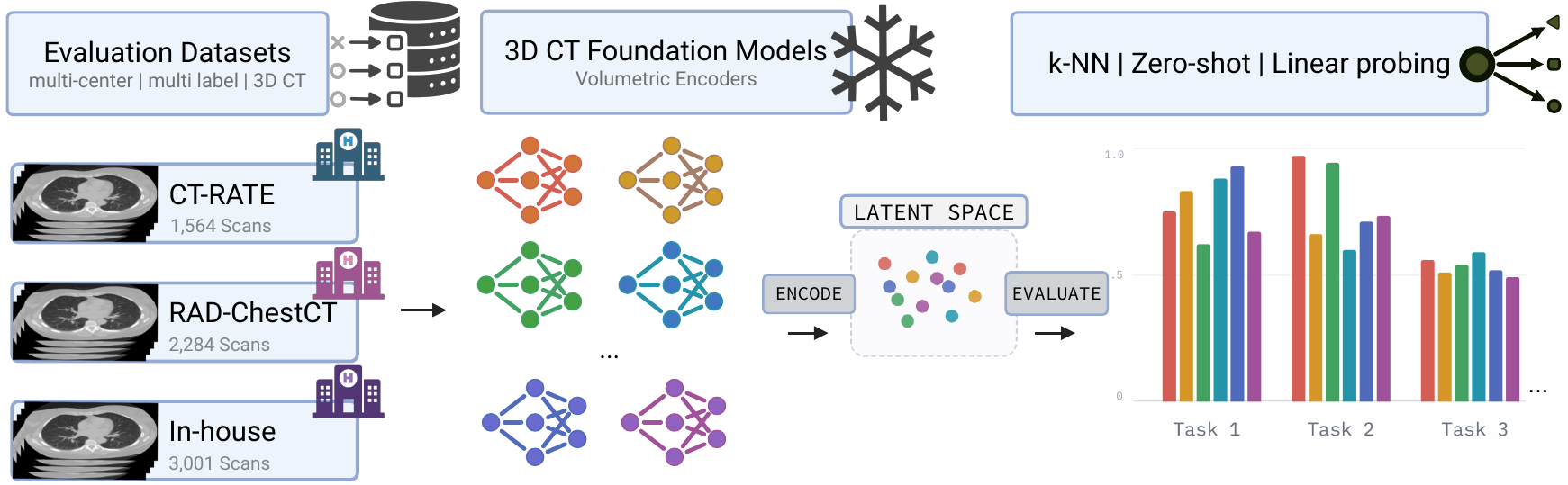}
\caption{\textbf{Evaluation Overview.} Three 3D thoracic CT datasets are encoded using frozen foundation models. The extracted representations are then evaluated across multiple findings classification tasks using $k$-NN, zero-shot classification, and linear probing readouts.}

\label{fig:pipeline_overview}
\end{figure}

Routine CT interpretation is inherently comprehensive: Regardless of the primary indication, radiologists inspect the entire volume \cite{herring2019learning,aydin2025emergency} and report any anatomical deviations, including incidental findings \cite{evans2022incidental}. 3D CT foundation models (FMs) can assist with this comprehensive review because they encode broad, generalizable representations of anatomy and pathology.
This generalizability is critical because medical pathologies follow a heavy-tailed distribution \cite{zhang2023deep}, making it infeasible to train dedicated classifiers for every possible abnormality. Furthermore, while hospitals may possess the infrastructure to run inference locally, they rarely have the labeled datasets necessary to fine-tune task heads for hundreds of rare conditions \cite{hoelzle2026longitudinal}. 
While fine-tuning remains a practical route for targeted applications, a true foundation model should, to a meaningful degree, inherently capture the diverse abnormalities present in a CT within its embeddings. Assessing whether current models achieve this fundamental capability requires isolating the latent space and benchmarking their frozen representations directly.
We therefore ask: \emph{Are current CT foundation models capable of incidental-finding detection without any fine-tuning, does a consistently best-performing model emerge, and where do their representational limitations lie?}

To answer this, we evaluate \emph{ten} 3D CT encoders in their frozen state on \emph{three} thoracic CT datasets: CT-RATE \cite{hamamci2024ctrate}, RAD-ChestCT \cite{draelos2021radchest}, and an in-house hospital cohort guaranteed to be unseen by any model.
To assess models' utility under data and compute scarcity, we avoid fine-tuning entirely and evaluate them solely through zero-shot prompting, $k$-nearest-neighbour classification, and linear probing. 
While no single model wins across all readouts, architectures utilizing fine-grained image tokenizers combined with report-aligned pre-training consistently lead performance. Furthermore, we observe that diagnostic difficulty is governed not by the choice of encoder, but by characteristics of the findings like its spatial extent and contrast, leaving small focal lesions challenging for all encoders.
Ultimately, this benchmark offers independent insights to inform the selection of baseline encoders and highlights critical representational bottlenecks for future model development.

\section{Results}

\begin{figure}[t]
    \centering
    \includegraphics[width=0.9\textwidth]{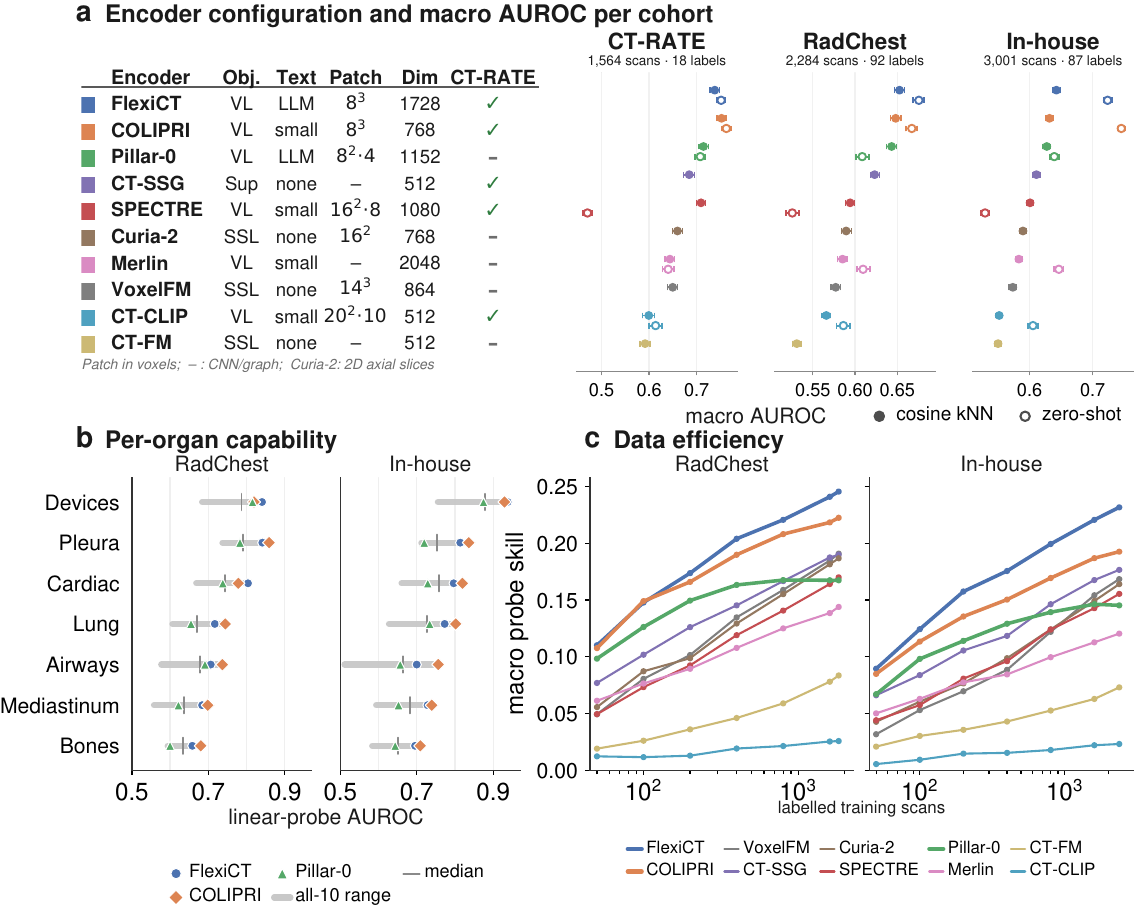}
\caption{\textbf{Benchmarking out-of-the-box capability and data efficiency.} 
\textbf{(a)}~Encoder configurations and macro AUROC. Models combining vision--language alignment with fine-grained tokenizers (FlexiCT, COLIPRI) consistently lead $k$NN and zero-shot performance across both public and unseen external cohorts. 
\textbf{(b)}~Per-organ linear probing confirms this architectural advantage persists across anatomical regions, though absolute detectability remains organ-dependent. 
\textbf{(c)}~Data efficiency curves demonstrate that these leading representations scale log-linearly with available data. Surprisingly, despite its strong overall performance, Pillar-0 rapidly hits a representational ceiling and saturates. Confidence intervals are omitted for readability as they are negligibly small.}
\label{fig:which-is-best}
\end{figure}

We evaluate ten 3D CT FMs as shown in the Fig.~\ref{fig:which-is-best}a, out of these, three models (CT-FM~\cite{pai2025ctfm}, Curia-2~\cite{saporta2026curia2}, VoxelFM~\cite{morenoaguado2026voxelfm}) use image-only pre-training, CT-SSG \cite{dipiazza2026ctssg} is supervised trained, while the remainder (CT-CLIP~\cite{hamamci2024ctrate}, COLIPRI~\cite{wald2026colipri}, SPECTRE~\cite{claessens2025spectre}, Merlin~\cite{blankemeier2024merlin}, Pillar-0~\cite{agrawal2025pillar0}, FlexiCT~\cite{li2026flexict}) use vision-language alignment. Five of these models used the CT-RATE training set during (pre)training. We evaluate the models using AUROC and prevalence-normalized PR-AUC-, also know as \emph{skill}: for a finding of prevalence $\pi$,
$\mathrm{skill}=(\mathrm{AP}-\pi)/(1-\pi)$, where $\mathrm{AP}$ is the average
precision, skill values range from $0$ to $1$ being the perfect score.


Our evaluation yields no single optimal model; rankings depend on how the embeddings are read (Fig.~\ref{fig:which-is-best}b, ~\ref{fig:which-is-best}c). Under cosine $k$NN, the top three encoders on every cohort are FlexiCT, COLIPRI, and Pillar-0, which are the transformers with the finest image tokenizer ($8$-voxel patches; Fig.~\ref{fig:which-is-best}a), whereas the coarsest-tokenizer transformer (CT-CLIP, $20{\times}20{\times}10$) ranks last. This patch-granularity advantage strongly co-occurs with report-aligned pretraining, demonstrating that fine-grained image tokenization combined with vision--language alignment generally yields the strongest out-of-the-box representations for retrieval. Crucially, Pillar-0 never saw CT-RATE yet remains in the top three on the two held-out cohorts, indicating that these shared architectural traits confer genuine generalization. However, data efficiency curves (Fig.~\ref{fig:which-is-best}c) reveal that  Pillar-0 rapidly hits a representational ceiling and saturates as labeled data increases, eventually being overtaken by CT-SSG. As the only \emph{supervised} encoder and the smallest embedding ($512$-d), CT-SSG stays highly competitive across both $k$NN and linear probing on all three cohorts. It matches or beats every self-supervised encoder and both convolutional models, suggesting that explicit labels can effectively substitute for scale. Under linear probing, COLIPRI maximizes AUROC and FlexiCT maximizes skill, at which point the strict patch-granularity ordering dissolves and the self-supervised VoxelFM rises to the front. CT-CLIP and CT-FM consistently underperform.

\begin{figure}[t]
    \centering
    \includegraphics[width=0.8\textwidth]{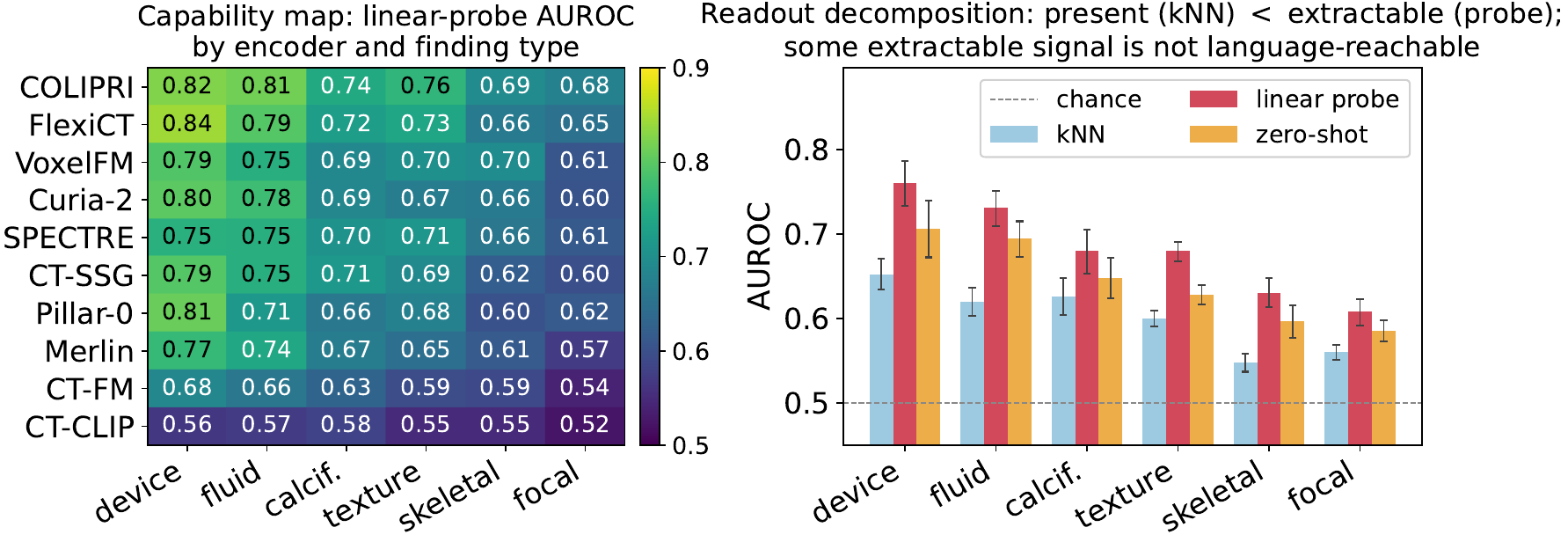}
\caption{\textbf{Classification difficulty is dictated by finding type, not encoder.} 
\textbf{Left:}~Linear-probe AUROC by model and finding type. The easy-to-hard hierarchy (columns, left to right) is highly consistent across all encoders: devices and fluid are universally accessible, while focal lesions remain challenging. 
\textbf{Right:}~Readout decomposition averaged across the six report-aligned encoders. Although linear probing extracts more latent signal than zero-shot prompting or $k$-NN retrieval, the underlying difficulty hierarchy is preserved. Focal abnormalities yield the lowest performance across all readouts.}
\label{fig:cap_readout}
\end{figure}

Performance differences across distinct findings are substantially larger than the differences between models on any given finding.
We sort every label into six phenotypic classes defined by visual morphology: calcification, devices, fluid, texture, focal, and skeletal (see Methods).
The easy-to-hard ordering across these types barely moves from one encoder to the next (Kendall $W=0.89$, Fig.~\ref{fig:cap_readout}a): each encoder reads support devices off easily, handles diffuse fluid and texture changes moderately, and struggles most with focal lesions. This limitation is representational rather than an artefact of the readout. The linear probe extracts more signal than $k$NN or zero-shot, yet focal performance stays low even under the probe (Fig.~\ref{fig:cap_readout}b). Moreover, it is selective \emph{within} the focal class: larger solid lesions stay retrievable (solid nodule skill value $0.26$, mediastinal lymphadenopathy $0.29$) while small subsolid ones sink toward chance (ground-glass nodule skill $0.05$, part-solid nodule $0.02$).

\begin{figure}[t]
    \centering
    \includegraphics[width=0.7\textwidth]{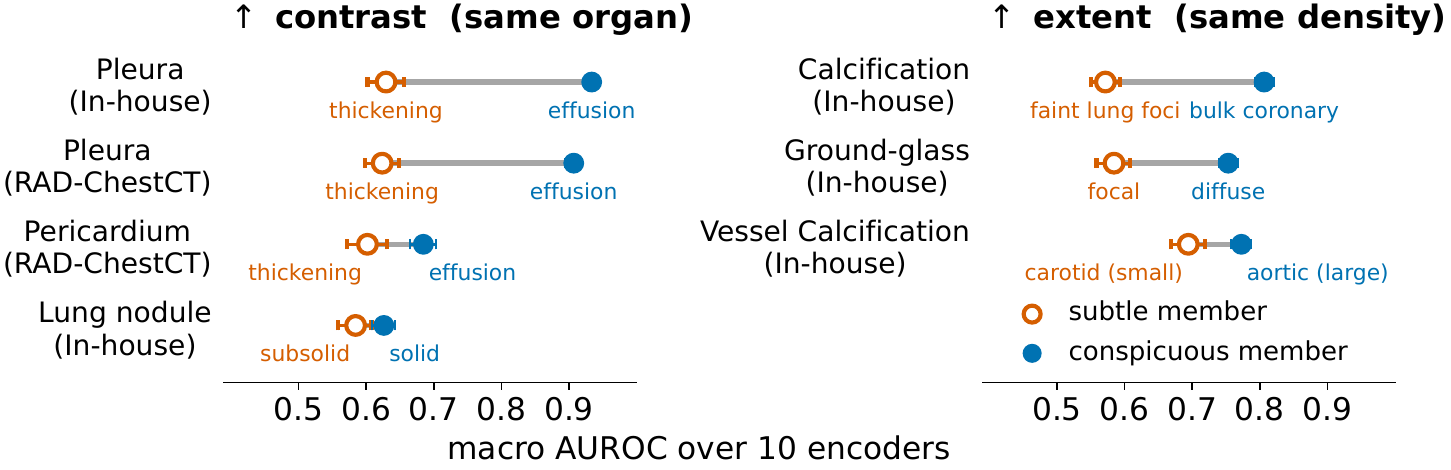}
\caption{\textbf{Contrast and spatial extent govern detectability.} Controlled within-organ comparisons demonstrate that increasing either contrast \textbf{(left)} or spatial extent \textbf{(right)} improves detection. Conspicuous findings (blue) systematically yield higher AUROCs than their subtle counterparts (orange) across all evaluated models. This indicates that a finding's physical footprint, not model architecture, is the primary bottleneck for incidental finding detection.}
\label{fig:contrast_extent}
\end{figure}

This difficulty ordering is driven by a straightforward empirical pattern: a finding's detectability scales with its \emph{contrast} against surrounding tissue and its spatial \emph{extent}. To isolate these physical effects from prevalence and anatomy, we conducted controlled within-organ comparisons (Fig.~\ref{fig:contrast_extent}). Holding extent fixed while increasing contrast (e.g., pleural and pericardial effusion vs.\ thickening; solid vs.\ subsolid nodule) improves macro AUROC by +0.18 on average. Conversely, holding contrast fixed and increasing extent (e.g., bulk vs.\ faint calcification; diffuse vs.\ focal ground-glass) yields a +0.16 improvement. All seven within-organ comparisons are positive and bootstrap-significant, with the ten evaluated encoders demonstrating near-universal agreement on this directionality (68 of 70 comparison $\times$ encoder pair). This pattern replicates robustly across both the public and unseen internal cohorts. Consequently, we observe that the representational bottleneck is primarily physical: low-contrast, small-extent abnormalities, which are precisely the subsolid focal lesions highlighted above, remain fundamentally challenging for all globally pooled encoders.

\begin{figure}[t]

\centering\includegraphics[width=\textwidth]{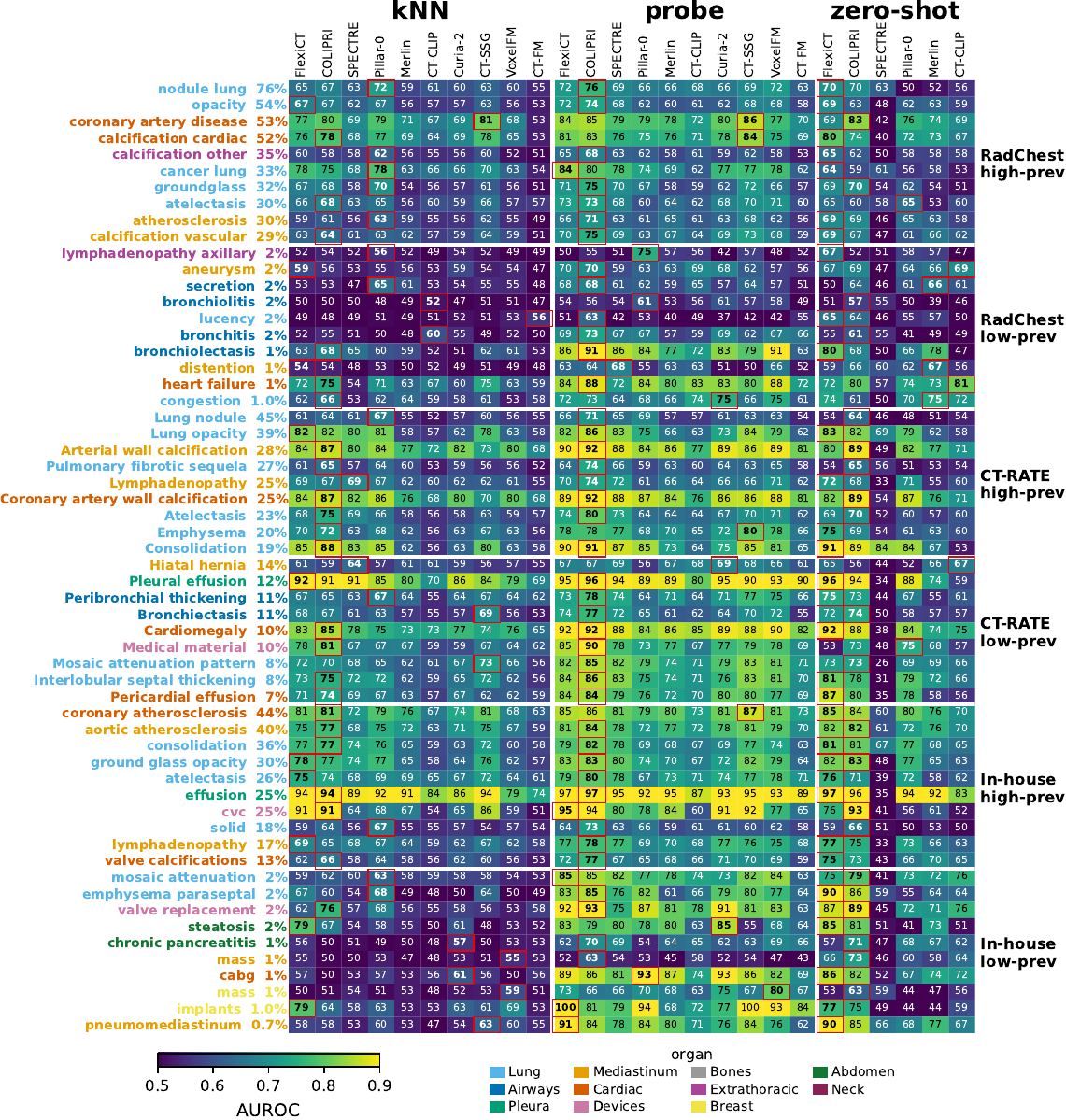}
\caption{\textbf{Comprehensive performance landscape across findings, cohorts, and readouts.} Per-finding AUROC for high- and low-prevalence abnormalities across three datasets, evaluated via $k$NN, linear probing, and zero-shot prompting. The prominent horizontal banding confirms that diagnostic difficulty is inherently tied to the finding itself, rather than the encoder. While linear probing consistently maximizes extractable signal, rare (low-prevalence) and focal findings remain universally challenging across all models and readout strategies. Conversely, structural abnormalities like support devices and fluid collections are robustly detected by nearly every encoder. Furthermore, the vertical banding reveals that while fine-grained tokenizers drive peak performance among the leading models, coarser architectures like CT-CLIP systematically collapse toward chance. The persistence of these difficulty patterns in the bottom block, our unseen internal hospital cohort, demonstrates that these representational bottlenecks are fundamental to the models themselves, rather than artifacts of the public pretraining datasets.}
\label{fig:supp_perclass}
\end{figure}

Finally, read-out performance varies across the finding prevalence range. The linear probe remains the strongest readout overall. For the rarest findings, zero-shot prompting yields a paired advantage of approximately $0.10$ AUROC over $k$NN (Fig.~\ref{fig:supp_perclass}). However, this advantage disappears when evaluated using prevalence-normalized skill. Under the skill metric, both zero-shot and $k$NN readouts score near zero on the rarest findings.

\section{Discussion and Outlook}
We investigated whether the frozen embeddings of current 3D CT foundation models possess the representational breadth necessary to capture the wide array of incidental findings encountered in routine interpretation.
Our results show that no single encoder consistently wins across all cohorts and readouts. Instead, the fundamental factor that transfers across cohorts is a physical bottleneck. Specifically, a finding's detectability is heavily influenced by its contrast and spatial extent. This leaves low-contrast, small-extent findings, such as subsolid focal lesions, universally challenging for globally pooled embeddings, approaching chance performance even under linear probing (ground-glass nodule skill value $0.05$, part-solid nodule $0.02$).

This physical limitation clarifies and extends recent evaluations of frozen foundation models. While prior work demonstrates that frozen features support in-distribution oncology diagnosis but falter on prognosis~\cite{aertspai2025fmbiomarkers}, our benchmarking across a wide spectrum of incidental findings elucidates why frozen embeddings systematically fail to capture certain pathologies. A globally pooled embedding inherently attenuates the subtle spatial signals of small lesions. In contrast, purpose-built systems that rely on explicit spatial localization \cite{mikhael2023sybil,brandt2026lungevaty} and per-structure radiomics achieve strong performance on incidental findings~\cite{marcinkiewicz2024incidental}, underscoring the necessity of spatial grounding.

We acknowledge several limitations: evaluating frozen embeddings bounds achievable peak performance, and the NLP-based extraction of labels across all three cohorts introduces noise that disproportionately affects the rarest findings. However, the difficulty ordering by finding type is highly consistent across all three cohorts (Kendall W=0.89), despite each using an independently constructed label pipeline. This strong agreement makes label noise an unlikely cause of the focal-lesion failure, indicating the pattern is representational.

These findings provide concrete directions for future model development. Because diagnostic difficulty is governed by contrast and extent rather than model capacity, overcoming this will likely require region- or lesion-level pretraining objectives that explicitly preserve small, low-contrast structures. Furthermore, the performance gap between linear probing and zero-shot retrieval suggests that relevant visual signals are present in the embeddings but remain inaccessible to the text encoder. Improving fine-grained vision--language alignment, perhaps coupled with lightweight, pattern-specific detection heads, offers a more promising route to robust clinical translation than relying on a single global vector.

\section{Data and Method}

\noindent\textbf{Datasets.} We evaluate on three distinct chest CT cohorts. The CT-RATE~\cite{hamamci2024ctrate} validation set initially contains 3{,}039 thoracic CT volumes from 1{,}304 patients. To prevent leakage from near-duplicate reconstructions, we filter this to one reconstruction per study, yielding 1{,}564 scans. This split is strictly disjoint from the CT-RATE training data seen by five of the evaluated models. The RadChestCT~\cite{draelos2021radchest} validation set comprises 2{,}284 single-scan patients. Our unseen internal hospital cohort includes 3{,}001 scans from 2{,}087 patients. Abnormality labels span the thoracic field of view and are derived from radiology reports: CT-RATE (18 labels) uses a fine-tuned text model, while the internal cohort (87 labels) utilizes the RATE framework~\cite{agrawal2025pillar0}. RadChestCT uses a radiologist-validated rule-based extractor; we aggregate its fine-grained location annotations into 92 labels by pooling across anatomical regions, preserving regional splits only for inherently multi-organ findings (cancer, mass, lymphadenopathy, calcification). Because patients in CT-RATE and the internal cohort contribute multiple scans, we evaluate both using strictly patient-grouped cross-validation.

\noindent\textbf{Models.} We evaluate 10 models capable of multi-abnormality classification. With the exception of the supervised CT-SSG~\cite{dipiazza2026ctssg}, all models utilize self-supervised learning: either image-only (CT-FM~\cite{pai2025ctfm}, Curia-2~\cite{saporta2026curia2}, VoxelFM~\cite{morenoaguado2026voxelfm}) or vision--language alignment (CT-CLIP~\cite{hamamci2024ctrate}, COLIPRI~\cite{wald2026colipri}, SPECTRE~\cite{claessens2025spectre}, Merlin~\cite{blankemeier2024merlin}, Pillar-0~\cite{agrawal2025pillar0}, FlexiCT~\cite{li2026flexict}). Pretraining data varies significantly across the lineup. Five models were exposed to the CT-RATE dataset during pretraining (indicated in Fig.~\ref{fig:which-is-best}). Furthermore, anatomical exposure ranges from strictly chest-specific (COLIPRI, CT-CLIP, CT-SSG, Pillar-0) and thoraco-abdominal (SPECTRE), to broad multi-region or whole-body collections (FlexiCT, VoxelFM, Curia-2, CT-FM), or exclusively abdomen--pelvis without chest exposure (Merlin). Consequently, the domain overlap with our chest-dominated evaluation cohorts differs markedly among encoders. We evaluate each model strictly as a frozen feature extractor using its author-prescribed preprocessing.

\noindent\textbf{Evaluation.}  We read the frozen embeddings using three distinct methods: cosine-weighted $k$-nearest neighbours ($k=5$); zero-shot classification (for vision--language models); and linear probing. For zero-shot classification, we score each finding against a fixed prompt pair---``A chest CT scan showing \{finding\}.'' and ``A chest CT scan showing no \{finding\}.''---computing $\sigma(\cos(z,t^{+})-\cos(z,t^{-}))$ over the $\ell_2$-normalized image and text embeddings. The scan's own report is never used. 
The linear probe is a one-vs-rest $\ell_2$-regularized logistic regression ($C=1$, balanced classes) fit on the raw embeddings.
All three readouts are evaluated using the same five-fold cross-validation, strictly grouped by patient to prevent leakage. We report $95\%$ confidence intervals from $1{,}000$ paired patient-level bootstrap resamples, computing macro averages over findings with at least $20$ positives. 

\noindent\textbf{Finding Types.} To analyze representational capability by finding characteristics rather than individual label, we group all labels across the three cohorts into six broad radiological phenotypes: \textbf{calcification} (e.g., coronary deposits), \textbf{devices} (e.g., pacemakers, stents), \textbf{fluid} collections and morphometric changes (e.g., effusions, cardiomegaly), diffuse \textbf{texture} patterns (e.g., atelectasis, ground-glass attenuation), \textbf{focal} space-occupying lesions (e.g., nodules, tumours), and \textbf{skeletal} findings (e.g., fractures, arthritis). The complete list is available on our github repository. This taxonomy is applied uniformly across all cohorts, serving as our primary unit of analysis for capability mapping (Fig.~\ref{fig:cap_readout}a). We measure the consistency of difficulty ordering across encoders using Kendall's coefficient of concordance ($W$) on the resulting matrix. For the contrast~$\times$~extent analysis (Fig.~\ref{fig:contrast_extent}), we isolate these physical variables from prevalence and anatomy by constructing controlled within-organ comparisons: pairs of findings in the same organ that differ along a single axis---either contrast (e.g.\ effusion vs.\ thickening) or extent (e.g.\ diffuse vs.\ focal ground-glass). For each pair, we test the difference in macro AUROC (averaged across all ten encoders) using a patient-grouped paired bootstrap, reporting how many encoders agree on the direction of the effect.

\clearpage

\paragraph{\textbf{Acknowledgments.}}
We thank the members of the IDERHA consortium - https://www.iderha.org/.
This project is supported by the Innovative Health Initiative Joint Undertaking (IHI JU) under grant agreement No.\ 101112135. The JU receives support from the European Union's Horizon Europe research and innovation programme and COCIR, EFPIA, Europa Bío, MedTech Europe, and Vaccines Europe. Funded by the European Union, the private members, and those contributing partners of the IHI JU. Views and opinions expressed are those of the authors only and do not necessarily reflect those of the aforementioned parties; neither of the aforementioned parties can be held responsible for them.

\paragraph{\textbf{Compliance with Ethical Standards.}}
All procedures complied with the Declaration of Helsinki and relevant institutional guidelines. The retrospective analysis of the in-house dataset was approved by the Technical University of Munich Ethics Committee (87/18S), with a waiver of informed consent. Use of the external CT-RATE and RAD-Chest CT datasets was governed by their respective institutional approvals: the Istanbul Medipol University Clinical Research Ethics Committee (E-10840098-772.02-6841) and the Duke University Health System IRB. All data across the three datasets were fully anonymized, collected retrospectively, and processed under informed consent waivers and HIPAA compliance where applicable.

\paragraph{\textbf{Code availability.}}
Code to reproduce the frozen-encoder readouts (AUROC and prevalence-normalized
skill under $k$-NN, zero-shot, and linear probing) on the two public cohorts
(RAD-ChestCT, CT-RATE) is available at
\url{https://github.com/maulikchevli/frozen-lexpert}. Pretrained weights
are obtained from each model's public release; the restricted internal cohort is
not included.

\bibliographystyle{splncs04}   
\bibliography{refs}

\end{document}